\pdfoutput=1

\documentclass[11pt]{article}

\usepackage{acl}

\usepackage{times}
\usepackage{latexsym}
\usepackage{cleveref}

\usepackage[T1]{fontenc}

\usepackage[utf8]{inputenc}

\usepackage{microtype}

\usepackage{inconsolata}

\usepackage{graphicx}

\usepackage{booktabs}
\usepackage{float}
\usepackage{caption}
\usepackage{subcaption}

\usepackage{hyperref}

    \title{Less is More: Parameter-Efficient Selection \\of Intermediate Tasks for Transfer Learning}

\author{David Schulte, Felix Hamborg, Alan Akbik \\
  Humboldt University of Berlin \\
  \texttt{davidsiriusschulte@gmail.com} \\
  \texttt{\{felix.hamborg, alan.akbik\}@hu-berlin.de}\\
}

\begin{document}
\maketitle
\begin{abstract}

Intermediate task transfer learning can greatly improve model performance. If, for example, one has little training data for emotion detection, first fine-tuning a language model on a sentiment classification dataset may improve performance strongly. But which task to choose for transfer learning? Prior methods producing useful task rankings are infeasible for large source pools, as they require forward passes through all source language models. We overcome this by introducing Embedding Space Maps (ESMs), light-weight neural networks that approximate the effect of fine-tuning a language model. 
We conduct the largest study on NLP task transferability and task selection with 12k source-target pairs. We find that applying ESMs on a prior method reduces execution time and disk space usage by factors of 10 and 278, respectively, while retaining high selection performance (avg. regret@5 score of 2.95). 
\end{abstract}

\section{Introduction} \label{sec:1}

The current default approach for supervised learning in NLP involves directly fine-tuning a pre-trained transformer using labeled data of the target task. However, prior work showed that in some cases it is beneficial to perform two consecutive fine-tunings in a row: first, on an \textit{intermediate task}, and then on the target task~\cite{phang2018sentence, vu-etal-2020-exploring}. This may be particularly effective when little training data exists for the target task, while much exists for the intermediate task. 

However, whether and how much performance is gained with intermediate task transfer learning heavily depends on the chosen intermediate task. Worse, finding the best intermediate task for a given target task is a non-trivial problem given the large amount of labeled datasets that exist for NLP. For instance, the HuggingFace Hub alone contains more than 160k datasets and 700k models. This renders an exhaustive search for the best possible intermediary task infeasible. The problem of finding promising intermediate tasks for a target task is called intermediate task selection.\footnote{In contrast, source selection includes ranking source models of any kind, e.g., not only intermediate tasks (already fine-tuned language models), but also only pre-trained models.}



Prior work investigates approaches for finding suitable intermediate tasks given a source transformer LM and a target task~\cite{achille2019task2vec, bassignana-etal-2022-evidence, li2021ranking, nguyen2020leep, 
tran2019transferability, vu-etal-2020-exploring}. These approaches rely on the local availability of (large) source models or a space-intense representation of source datasets. 
Methods also require resource-intensive computation for each source-target pair \cite{poth-etal-2021-pre, you2022ranking}. 
Thus, most approaches are infeasible in real-world scenarios, i.e., with large source pools and constrained resources. While the large pool of available models and datasets is a valuable resource, user cannot optimally utilize it \cite{you2022ranking}.

This paper makes two contributions. 
First, we propose Embedding Space Maps (ESMs), linear transformations of the embedding space, to be used in combination with LogME \citep{you2021logme}, a source selection method that achieves high selection performance but suffers from its dependency on forward passes through each source model. We overcome this by approximating the embeddings of fine-tuned language models with ESMs.
The resulting source selection method ESM-LogME reduces execution time by a factor of 10 and disk space usage by a factor of 278, and thus enables efficient source selection also on large source pools.

Second, we compare the performance of ESM-LogME to prior methods in the to date largest study on transferability across NLP tasks with more than 1.5k source datasets from HuggingFace Hub and 8 target datasets across several task types and languages. The results show that ESM-LogME is the best-performing source selection method that is feasible in real-world scenarios. We release source code and all resources under the Apache 2 license. Our Python package also allows to share and find ESMs to facilitate efficient source selection among researchers and practitioners. The repository is available at: \href{https://github.com/davidschulte/hf-dataset-selector}{https://github.com/davidschulte/hf-dataset-selector}




\section{Related Work}\label{sec:2}
Transfer learning is a common paradigm in NLP.
With BERT, \citet{devlin-etal-2019-bert} propose encoders that are trained on a large corpus and then fine-tuned on individual target tasks.
\citet{phang2018sentence} show that language models can benefit from adding an intermediate fine-tuning step. This procedure is called intermediate task transfer learning. One of its challenges is finding the right intermediate task. 


Source selection methods determine transfer suitability of source tasks for a given target task. The resulting rankings enable users to pick the potentially best sources, e.g., to perform transfer learning using these top picks to find the actual best source. Methods typically consist of two resource-intense phases. In a one-time process, for each source a target-independent representation is created (P1). Then, for a given target task, a ranking is produced using these representations (P2).


TextEmb \cite{vu-etal-2020-exploring} and TaskEmb \cite{achille2019task2vec} embed datasets into a vector space and compute the distance of their representations. 
TaskEmb shows good performance in the literature, but its vector representation is in general as large as the language model itself. The vectors produced by TextEmb are small, but describe only the task domain and not the relation of inputs and labels.

NCE \cite{tran2019transferability}, LEEP \cite{nguyen2020leep}, and LogME \cite{you2021logme} rank source models by evaluating pseudo-labels, their distributions, and target embeddings.
These methods show state-of-the-art performance in source selection, but require forward passes through each source model. For scenarios with many source models, such approaches may thus be infeasible.

Prior studies evaluate the effect of intermediate task transfer learning and the performance of source selection methods on NLP tasks \cite{bassignana-etal-2022-evidence, poth-etal-2021-pre, vu-etal-2020-exploring}.
But these studies do not represent real-world scenarios. Employed source pools are small ($n<50$), whereas users can and have to choose from a very large pool of source datasets and models. Also, studies largely do not evaluate execution time and disk space usage. We argue that efficiency is crucial for practical use of source selection methods.

In sum, prior work can achieve high ranking accuracy, but does not explore source selection in real-world scenarios. Studies largely neglect efficiency and use benchmarks that do not resemble source pools available on popular model hubs.


\section{Embedding Space Maps}
Fine-tuning a base language model $f_0$ on a task $T$ using dataset $\mathcal{D}_T$ results in a fine-tuned language model $f_T$, which embeds texts differently than the base language model. We describe this effect as a function $h_{0 \rightarrow T}$ on the embedding space and approximate this function  using a neural network, which we call ESM, $\phi_{0 \rightarrow T}$.
Specifically, an ESM's inputs are embeddings produced by the base model $f_0$ and its outputs are $d$-dimensional approximations of the embeddings produced by the fine-tuned model $f_T$.
Attaching this network to a base model allows us to approximate how the respective fine-tuned model embeds a given text $x$.
\begin{equation}
    \phi_{0 \rightarrow T}(f_0(x)) \approx h_{0 \rightarrow T}(f_0(x)) = f_T(x)
\end{equation}

\begin{figure}[H]\centering\includegraphics[width=0.48\textwidth,height=\textheight,keepaspectratio]
{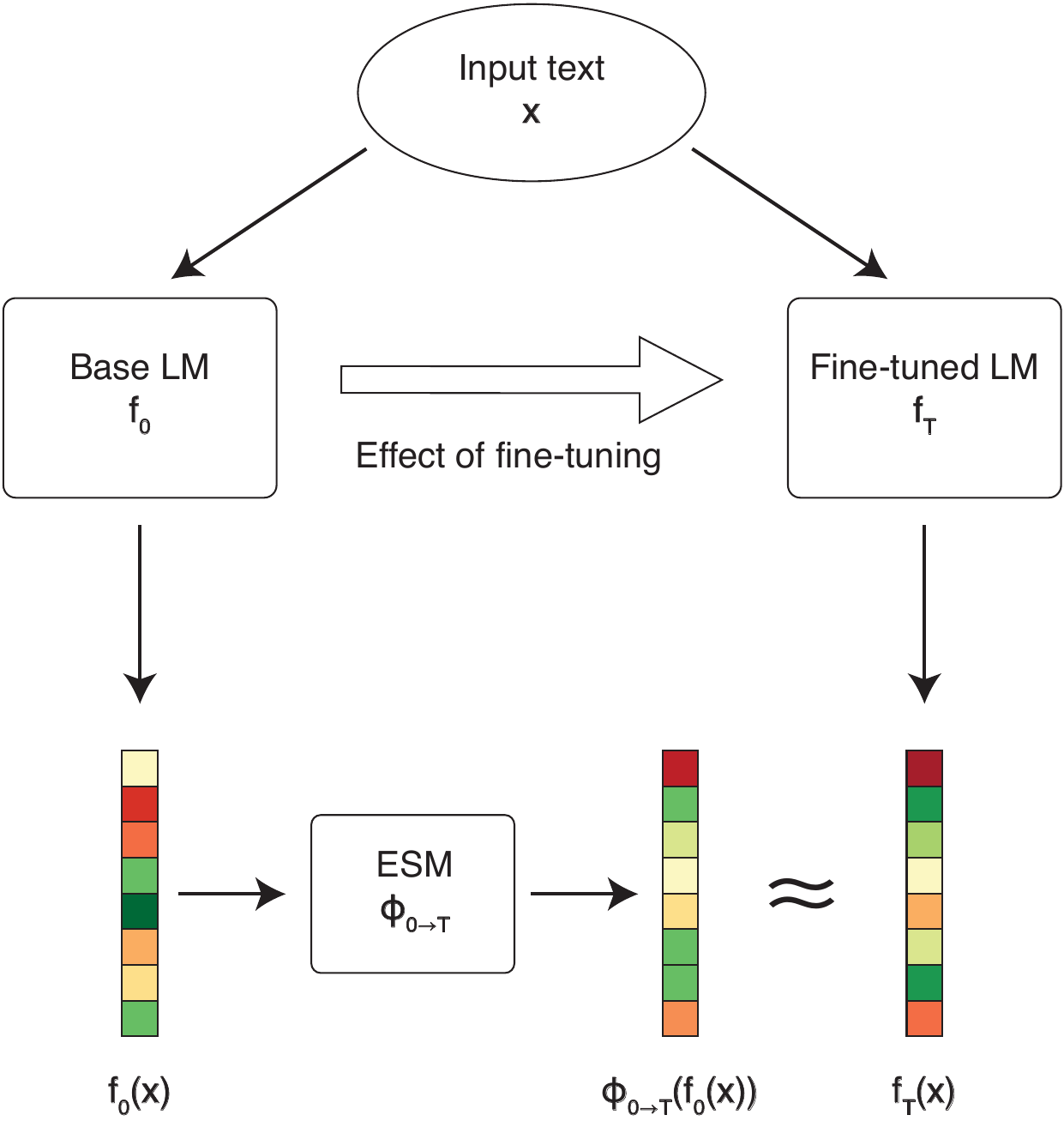}
    \caption{
    Embedding Space Maps approximate how a fine-tuned language model embeds an input text $x$ by transforming embeddings produced by the base model.}
    \label{fig:transfer-results}
\end{figure}

\begin{figure*}[!ht]
    \centering
    \begin{minipage}{0.317\textwidth}
        \centering
        \includegraphics[width=\linewidth]{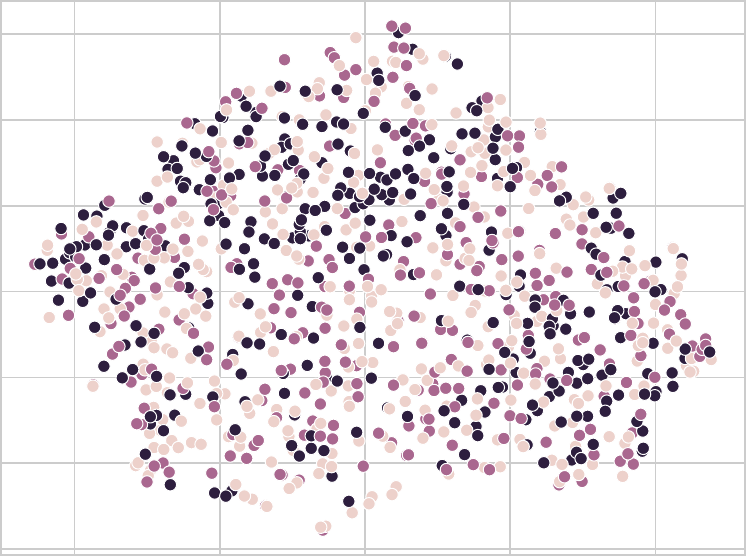}
    \end{minipage}%
    \hspace{0.25cm}
    \begin{minipage}{0.317\textwidth}
        \centering
        \includegraphics[width=\linewidth]{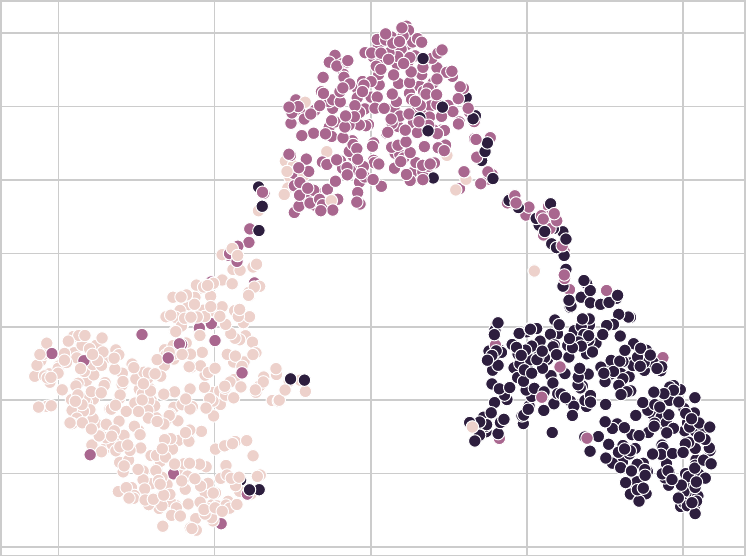}
    \end{minipage}
    \hspace{0.25cm}
    \begin{minipage}{0.317\textwidth}
        \centering
        \includegraphics[width=\linewidth]{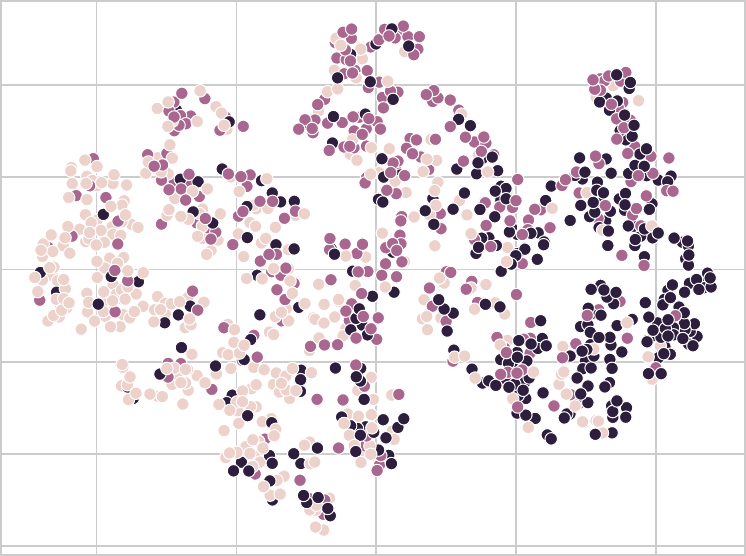}
    \end{minipage}\\
        \vspace{0.1cm}
    \begin{minipage}
    {0.5\textwidth}
        \centering
        \includegraphics[width=0.8\linewidth]{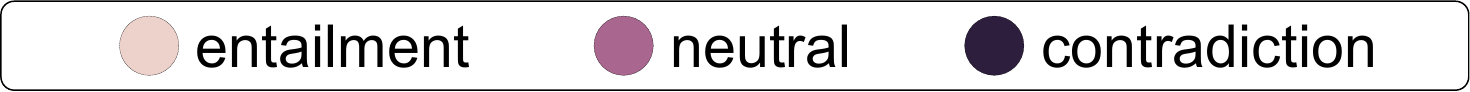}

    \end{minipage}
    \caption{
    We use T-SNE to visualize embeddings of inputs of the SNLI validation split using BERT (l.), BERT fine-tuned on SNLI (m.), and BERT and an ESM that was trained using the fine-tuned model (r.).
    The ESM-transformed embeddings are clearly arranged with regard to their classes. While classes are not as distinguished as when embedded by the fine-tuned model, a clear gradient is visible (albeit having applied dimension reduction).
    }
    \label{fig:t-sne-visualizations}
\end{figure*}
For phase P1 (cf. \Cref{sec:2}), i.e., to train an ESM, we embed each text $x$ of a dataset $\mathcal{D}$ with both $f_0$ and $f_T$. We train the ESM $\phi_{0 \rightarrow T}$ by using the resulting embeddings ($f_0(x), f_T(x)$) as train examples. Although any dataset could be used in this step, we choose to embed $D_T$ (the dataset that $f_0$ was fine-tuned on to obtain $f_T$), as its input texts describe task $T$ and its effect on the language model best.
\footnote{We train ESMs for 10 epochs with a learning rate of 0.001 and weight decay of 0.01.}


For P2, we once compute embeddings for the inputs of a target task using the base model and transform them using an ESM for each intermediate task. Following this, we rank sources by the LogME score of their ESM-transformed embeddings and target labels.
We call this workflow ESM-LogME. Since ESMs approximate the embeddings produced by the intermediate model, ESM-LogME can be viewed as an approximation of LogME.


\begin{table*}[tb]
\small

    \centering
\begin{tabular}{lrrrrrrrrrr}
\toprule
 & \multicolumn{4}{c}{Classification} & \multicolumn{4}{c}{Regression} & Runtime & Memory  \\
 & NDCG & R@1 & R@3 & R@5 & NDCG & R@1 & R@3 & R@5  & (ms) & (MB)\\ 
\midrule
ESM-LogME & 57 & 6.85 & 3.83 & 1.91 & 61 & 10.53 & 7.41 & 4.69 & \textbf{423} & 
\textbf{2}  \\
LogME & 82 & 2.89 & 0.12 & \textbf{0.12} & 86 & 1.64 & 1.64 & \textbf{1.64} & 4,501 & 639 \\
Vocabulary Overlap & 59 & 4.45 & 2.37 & 1.80 & 60 & 22.07 & 12.25 & 11.24 & 8,579 & 
\textbf{5} \\
TaskEmb & 46 & 15.28 & 13.62 & 13.08 & 80 & 7.38 & 3.02 & 3.02  & 2,767 & 639 \\
TextEmb & 54 & 7.97 & 7.26 & 6.73 & 54 & 7.32 & 11.52 & 11.10 & \textbf{0.01} & \textbf{0.01} \\
Frozen Transfer & 46 & 6.91 & 3.79 & 3.00 & 66 & 8.53 & 1.76 & \textbf{1.76}  & 10,541 & 639 \\
\midrule
NCE & 74 & 4.47 & 2.70 & 0.12 & \textbf{-} & \textbf{-} & \textbf{-} & \textbf{-} & 3,857 & 639 \\
LEEP & 82 & 1.90 & 0.12 & 0.12 & \textbf{-} & \textbf{-} & \textbf{-} & \textbf{-} & 3,893 & 639 \\
\bottomrule
\end{tabular}

\caption{Overview of Ranking Performances and Efficiency
}
    \label{tab:source_ranking_eval}
\end{table*}

We design ESMs as a single forward layer to minimize their size and computational complexity. Therefore, ESMs are linear transformations. This design choice greatly reduces the amount of parameters needed to describe fine-tuning $f_0$ on a task, e.g., from 110M to less than 0.6M for BERT. Thus, ESMs drastically reduce compute cost and disk space usage of source representations.

Since $h_{0 \rightarrow T}$ is the result of changing (many) parameters inside the language model, $\phi_{0 \rightarrow T}$ underfits this function. Our evaluation shows that---albeit their simplicity---ESMs can encode abstract characteristics of their corresponding task. For a more intuitive understanding of the concept of ESMs, we visualize how well they approximate the effect of fine-tuning in an experiment (see \Cref{fig:t-sne-visualizations}).

ESMs are parameter-efficient representations of transfer learning that are attached to a base language model. This modular design of ESMs resembles that of adapters \cite{houlsby2019parameter, pfeiffer-etal-2020-adapterhub}. While adapter blocks are inserted between transformer layers of a language model, ESMs are placed solely on top. Using adapters for source selection requires a forward pass through the entire language model for each source. This also holds for state-of-the-art methods such as Log-ME. In contrast, with ESMs, only a single forward pass through the base language model is required to compute the base embeddings of the target task. These can then quickly be transformed using an ESM for each intermediate task. In turn, ESMs significantly decrease computational effort in P2.




\section{Experimental Setup}
In contrast to prior studies, we aim to evaluate ranking performance in a real-world scenario. We parse datasets from the HuggingFace Hub and heuristically determine their input and label columns to gather as many intermediate tasks as possible. This process includes searching for common column names, analyzing column types and contents\footnote{Cf. \Cref{source-datasets-appendix}.}.
The resulting pool consists of 1553 datasets (1496 classification and 57 regression tasks).

We manually curate a selection of target datasets that is diverse as to task type, domain, and language. It consists of datasets or subsets from \textbf{IMDB} \cite{maas-etal-2011-learning}, TweetEval with Emotion and Sentiment subsets (\textbf{TES}, \textbf{TSS}) \cite{barbieri-etal-2020-tweeteval}, \textbf{J-STS} \cite{kurihara-etal-2022-jglue}, Multi-Dimensional Gender Bias Classification (\textbf{MDGB}) \cite{dinan-etal-2020-multi}, the English subset of \textbf{PAWS-X} \cite{yang-etal-2019-paws}, Query Wellformedness (\textbf{GQW}) \cite{faruqui-das-2018-identifying}, and Civil Comments (\textbf{GCC}) \cite{borkan2019nuanced}.\footnote{Cf. \Cref{target-datasets-appendix}.}
We artificially reduce the train size of target datasets to 1k rows to simulate data scarcity and of source datasets to 10k rows for evaluation efficiency.

We use BERT (\textit{bert-base-multilingual-uncased}), perform transfer learning for all source-target pairs\footnote{We train the model for 3 epochs with a learning rate of $2e-5$ and a weight decay of 0.01.}, and evaluate the rankings of several source selection methods using the realized performance gains on a validation dataset as ground truth.\footnote{Computations were run on a single Nvidia RTX 600 GPU.}
We calculate source rankings using ESM-LogME, LogME, NCE, LEEP, TextEmb, TaskEmb, vocabulary overlap (Jacard Index of the sets of tokenized inputs), and fine-tuning source models while freezing the parameters of the language model.
Model performance is measured in accuracy for classification tasks and as mean of Pearson correlation and Spearman's rank corr. coefficient for regression. 

We follow prior work \cite{vu-etal-2020-exploring, poth-etal-2021-pre} and measure the quality of source rankings using NDCG \cite{jarvelin2002cumulated} and regret@$k$ \cite{renggli2022model} with $k=1,3,5$ (all reported as pp.). Effectively, R@$k$ expresses how well the best task in the selected $k$ tasks performs compared to using the best task from the entire pool. The metric assumes that users employ transfer learning on all $k$ selected tasks to find the actual best from those. We use R@5 as the primary metric.


\section{Results}
\subsection{Transfer Results}
\Cref{fig:transfer-results} shows the performance for each target task distributed over all source tasks. With one exception, target tasks benefit from the majority of intermediate tasks, albeit to a different extent. Though, depending on the chosen intermediate task, transfer learning may also degrade performance compared to the base model.\footnote{The best sources generally are of the same task type, e.g., sentiment classification, as the target (cf. \Cref{target_top_10_tables}).} The findings highlight the effectiveness of intermediate task transfer learning, but also the importance of proper task selection.

\begin{figure}[tb]\centering\includegraphics[width=0.48\textwidth,height=\textheight,keepaspectratio]
{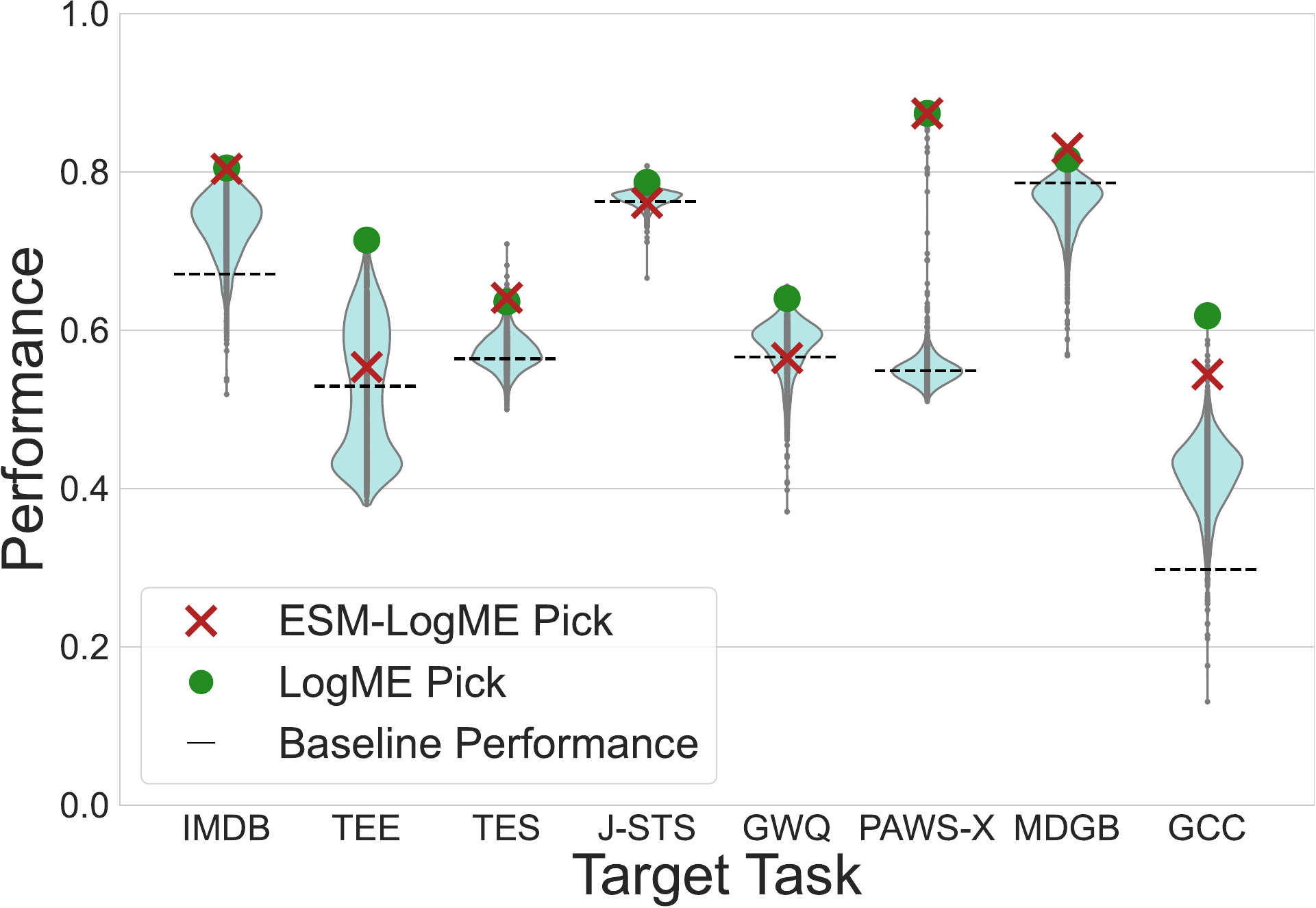}
    \caption{
    The baseline performance indicates the performance resulting from fine-tuning the base model without any intermediary task. Marks indicate the sources ranked highest by ESM-LogME and LogME.}
    \label{fig:transfer-results}
\end{figure}

\subsection{Source Ranking Evaluation} 



Model-based methods perform better than dataset-based approaches (\Cref{tab:source_ranking_eval}). In particular, LogME, NCE, and LEEP produce the best rankings (R@5 of 0.12 for classification and 1.64 for regression).\footnote{NCE and LEEP have to be treated separately: as they do not apply to regression tasks, we evaluate them on a source pool that contains only classification tasks.} ESM-LogME performs better than most remaining methods on classification target tasks (1.91). Its performance slightly worsens on regression targets (4.69). However, in 4 target tasks, the best source task is contained in the top 5 rankings of ESM-LogME (0). Averaged over all tasks, ESM-LogME yields R@5 of 2.95, i.e., transferring from the best of the top 5 picks leads to 97.05\% of the best possible performance of the entire source pool.\footnote{For detailed results per target task, cf. \Cref{source_ranking_details}.}

\Cref{fig:transfer-results} shows the source tasks ranked highest by ESM-LogME and LogME. While for 3 target tasks the top 1 picks of ESM-LogME lead to no significant transfer gains (or even a slight transfer loss), it also finds the best source for 2 of the target tasks (for 2 tasks, it even picks a better source task than LogME, likely due to a positive approximation error). This highlights that the rankings of ESM-LogME and other methods should be used to determine a small candidate set for transfer learning, rather than solely relying on the top pick.

\subsection{Efficiency}
Results from P1 are target-independent and can be shared publicly. We measure efficiency in P2, which needs to be performed by users for intermediate task selection.

\Cref{tab:source_ranking_eval} shows that ESM-LogME is by far the fastest and also most storage efficient selection method (aside from TextEmb, which yields inferior rankings). 
It is $\approx$10x faster than LogME, NCE, and LEEP and scales well across source tasks, since ESMs can quickly transform base embeddings. It is 278x more storage-efficient than model-based methods, which require trained source models. 




\section{Conclusion}
We show that a linear transformation of the embedding space suffices to describe a source task well enough for source selection. Although ESM-LogME yields less accurate rankings than LogME, our results show that the ESM-LogME workflow performs well on most target tasks (avg. regret@5 score of 2.95). At the same time, ESM-LogME is substantially more efficient than all well-performing state-of-the-art methods. This makes it the best-performing source selection method that is feasible in a real-world scenario.

\section{Limitations}
This study has the following limitations that result either from weaknesses of ESM-LogME or resource scarcity during the evaluation.


\subsection{Specificity to a language model}
One drawback of ESMs is that they are specific to a base language model. In practice, a user might not care which base model they use, but may care only about the performance on the target task. In this case, the user would have to compute target embeddings with several base language models and then apply ESM-LogME using ESMs specific to the corresponding language models.

\subsection{ESM Architectures}
We designed ESMs as single linear layers for simplicity and efficiency. However, other architectures are also worth exploring. We expect non-linear transformations to better approximate the effect of fine-tuning. On the downside, they are larger and have a higher risk of overfitting.

\subsection{Results across language models}
In this study, we evaluated ESM-LogME solely on a single base language model. As mentioned previously, users might want to consider several base models. Furthermore, we did not study how well the rankings of ESM-LogME specific to language model \textit{A} could be transferred to intermediate task transfer learning using language model \textit{B}. \citet{poth-etal-2021-pre} show a strong correlation of the performance of source-target pairs using the base models BERT and RoBERTa. This result indicates that intermediate task selection across language models may be viable.

\subsection{Dataset sizes}
To enable an evaluation across many source tasks, we considered only one configuration of dataset sizes, i.e., 10k source rows (or less) and 1k target rows. Prior work shows that the size of target datasets significantly affects transfer gains \cite{poth-etal-2021-pre, vu-etal-2020-exploring}. We did not research the effect of dataset sizes on task transferability and source selection accuracy.

\subsection{Dataset Selection}
Our heuristic parsing of the source datasets yields non-sensical datasets resulting from wrong assignments of input or label columns. It also forgoes many datasets whose columns it cannot assign. Additionally, it considers only one input-label-combination, whereas certain datasets have multiple combinations describing different tasks.
Although the target datasets are curated to be as diverse as possible, they only contain a single non-English-language dataset, i.e., J-STS. Thus, the dataset selection does not allow us to analyze the effect of transfer learning across languages. Though, our results on J-STS imply that language may have a significant affect on transferability, since the magnitude of transfer gains are low, and the best sources are largly defined by language and not task type. 

\subsection{Real-world applicability}
The assumption behind the efficiency of ESM-LogME is that users do not have to train the ESM for each source themselves. ESM-LogME is useful only if users can access ESMs specific to their chosen base language model and the source datasets in their source pool. This can be facilitated by either storing them on model hubs, such as HuggingFace, or by creating a model hub specific to ESMs, similar to Adapter Hub \cite{pfeiffer-etal-2020-adapterhub}. We plan to initially publish ESMs created for the paper at hand and currently investigate both options.

\section*{Acknowledgments}
We thank all reviewers for their valuable com-
ments. Felix Hamborg was supported by the WIN program of the Heidelberg Academy of Sciences and Humanities, financed by the Ministry of Science, Research and Arts of the State of Baden-Wurttemberg,
Germany, and by the State Criminal Police Office of the State of North Rhine-Westphalia, Germany.


\bibliography{custom}

\newpage
\appendix

\clearpage
\onecolumn

\section{Detailed Results}

\subsection{Top 10 Source Tasks and ESM-LogME Picks}\label{target_top_10_tables}

\begin{table}[H]
    \centering
    \begin{tabular}{cc} 

    \begin{minipage}{0.5\textwidth}
        \centering
        \scriptsize
        \begin{tabular}{l p{4cm} rr}
\toprule
 & Source Task & Perf. & ESM-LM Rank \\
\midrule
1 & rotten\_tomatoes:default & 81.0 & 9 \\
2 & amazon\_polarity:amazon\_polarity & 80.5 & 6 \\
3 & sst:dictionary & 80.4 & 1 \\
4 & yelp\_polarity:plain\_text & 80.3 & 17 \\
5 & senti\_lex:hi & 80.3 & 823 \\
6 & BDas/EnglishNLPDataset:EnglishData & 80.2 & 43 \\
7 & senti\_lex:bg & 80.2 & 204 \\
8 & KBLab/overlim:sst\_da & 80.1 & 83 \\
9 & tweet\_eval:emotion & 80.0 & 161 \\
10 & silicone:sem & 80.0 & 679 \\
\bottomrule
\end{tabular}

        \caption{Ground Truth Ranking: IMDB}
        \label{fig:prob1_6_2}
        \vspace{1cm}
    \end{minipage}%
    \begin{minipage}{0.5\textwidth}
        \centering    
        \scriptsize
\begin{tabular}{l p{4cm} rr}
\toprule
 & Source Task & Perf. & True Rank \\
\midrule
1 & sst:dictionary & 80.4 & 3 \\
2 & sst:default & 78.6 & 65 \\
3 & kuroneko5943/snap21:CDs\_and\_Vinyl\_5 & 78.6 & 63 \\
4 & kuroneko5943/snap21:Video\_Games\_5 & 77.4 & 172 \\
5 & kuroneko5943/snap21:Movies\_and\_TV\_5 & 79.3 & 30 \\
6 & amazon\_polarity:amazon\_polarity & 80.5 & 2 \\
7 & glue:sst2 & 79.9 & 16 \\
8 & Patt/ReCoRD\_TH\_drop:default & 72.2 & 1035 \\
9 & rotten\_tomatoes:default & 81.0 & 1 \\
10 & evaluate/glue-ci:sst2 & 79.9 & 15 \\
\bottomrule
\end{tabular}

        \caption{ESM-LogME Ranking: IMDB}
        \label{fig:prob1_6_1}
            \vspace{1cm}
    \end{minipage}
    \end{tabular}
    \vspace{1cm}
    \begin{tabular}{cc}
    \begin{minipage}{0.5\textwidth}
        \centering
        \scriptsize
        \begin{tabular}{l p{4cm} rr}
\toprule
 & Source Task & Perf. & ESM-LM Rank \\
\midrule
1 & emo:emo2019 & 72.73 & 287 \\
2 & tasksource/crowdflower:text\_emotion & 71.39 & 173 \\
3 & silicone:meld\_s & 71.39 & 1109 \\
4 & tyqiangz/multilingual-sentiments:ind\dots & 70.86 & 1054 \\
5 & Deysi/sentences-and-emotions:default & 69.79 & 357 \\
6 & Sharathhebbar24/app\_reviews\_modded\dots & 69.79 & 674 \\
7 & scaredmeow/shopee-reviews-tl-bin\dots & 69.52 & 1156 \\
8 & tyqiangz/multilingual-sentiments\dots & 69.25 & 985 \\
9 & tyqiangz/multilingual-sentiments:all & 68.98 & 1277 \\
10 & tweet\_eval:sentiment & 68.98 & 128 \\
\bottomrule
\end{tabular}

        \caption{Ground Truth Ranking: TEE}
        \label{fig:prob1_6_2}
    \end{minipage}%
    \begin{minipage}{0.5\textwidth}
        \centering    
        \scriptsize
\begin{tabular}{l p{4cm} rr}
\toprule
 & Source Task & Perf. & True Rank \\
\midrule
1 & sst:dictionary & 55.35 & 643 \\
2 & philschmid/emotion:split & 66.31 & 52 \\
3 & Patt/ReCoRD\_TH\_drop:default & 42.78 & 1323 \\
4 & sst:default & 54.28 & 690 \\
5 & dair-ai/emotion:split & 66.31 & 45 \\
6 & google/civil\_comments:default & 63.64 & 148 \\
7 & dair-ai/emotion:unsplit & 66.58 & 40 \\
8 & ttxy/emotion:default & 66.31 & 47 \\
9 & d0rj/rudetoxifier\_data:default & 59.63 & 372 \\
10 & sst2:default & 63.37 & 160 \\
\bottomrule
\end{tabular}

        \caption{ESM-LogME Ranking: TEE}
        \label{fig:prob1_6_1}
    \end{minipage}
    \end{tabular}
    \vspace{1cm}
    \begin{tabular}{cc}
    \begin{minipage}{0.5\textwidth}
        \centering
        \scriptsize
        \begin{tabular}{l p{4cm} rr}
\toprule
 & Source Task & Perf. & ESM-LM Rank \\
\midrule
1 & cardiffnlp/tweet\_sentiment\_multilingual:all & 70.9 & 4 \\
2 & cardiffnlp/tweet\_sentiment\_multilingual:e\dots & 68.2 & 51 \\
3 & tyqiangz/multilingual-sentiments:eng\dots& 66.8 & 36 \\
4 & BDas/EnglishNLPDataset:EnglishData & 65.8 & 29 \\
5 & MichiganNLP/TID-8:goemotions-ann & 65.3 & 44 \\
6 & Areeb123/drug\_reviews:default & 64.2 & 22 \\
7 & sst:default & 64.1 & 3 \\
8 & tasksource/crowdflower:airline-sent\dots & 64.1 & 1 \\
9 & MichiganNLP/TID-8:sentiment-atr & 63.8 & 14 \\
10 & MichiganNLP/TID-8:goemotions-atr & 63.6 & 16 \\
\bottomrule
\end{tabular}

        \caption{Ground Truth Ranking: TES}
        \label{fig:prob1_6_2}
    \end{minipage}%
    \begin{minipage}{0.5\textwidth}
        \centering    
        \scriptsize
\begin{tabular}{l p{4cm} rr}
\toprule
 & Source Task & Perf. & True Rank \\
\midrule
1 & tasksource/crowdflower:airline-sent\dots & 64.1 & 8 \\
2 & sst:dictionary & 62.5 & 32 \\
3 & sst:default & 64.1 & 7 \\
4 & cardiffnlp/tweet\_sentiment\_multilingual:all & 70.9 & 1 \\
5 & tasksource/crowdflower:text\_emotion & 62.6 & 29 \\
6 & yelp\_polarity:plain\_text & 62.6 & 26 \\
7 & tweet\_eval:offensive & 61.5 & 53 \\
8 & MichiganNLP/TID-8:sentiment-ann & 62.8 & 22 \\
9 & claritylab/utcd:out-of-domain & 55.7 & 1199 \\
10 & glue:sst2 & 62.9 & 19 \\
\bottomrule
\end{tabular}

        \caption{ESM-LogME Ranking: TES}
        \label{fig:prob1_6_1}
    \end{minipage}
    \end{tabular}
    \vspace{1cm}
    \begin{tabular}{cc}
        \centering
    \begin{minipage}{0.5\textwidth}
        \centering
        \scriptsize
        \begin{tabular}{l p{4cm} rr}
\toprule
 & Source Task & Perf. & ESM-LM Rank \\
\midrule
1 & llm-book/JGLUE:JNLI & 80.76 & 653 \\
2 & shunk031/JGLUE:JNLI & 80.76 & 1006 \\
3 & clue:cmnli & 79.38 & 665 \\
4 & shunk031/jsnli:without-filtering & 79.27 & 888 \\
5 & shunk031/jsnli:with-filtering & 79.14 & 998 \\
6 & xtreme:XNLI & 78.99 & 1060 \\
7 & PNLPhub/FarsTail:FarsTail & 78.99 & 939 \\
8 & paws:labeled\_final & 78.86 & 364 \\
9 & csebuetnlp/xnli\_bn:xnli\_bn & 78.77 & 910 \\
10 & stsb\_multi\_mt:zh & 78.67 & 664 \\
\bottomrule
\end{tabular}

        \caption{Ground Truth Ranking: J-STS}
        \label{fig:prob1_6_2}
    \end{minipage}%
    \begin{minipage}{0.5\textwidth}
        \centering    
        \scriptsize
\begin{tabular}{l p{4cm} rr}
\toprule
 & Source Task & Perf. & True Rank \\
\midrule
1 & kejian/codeparrot-train-more-filter\dots & 76.1 & 1188 \\
2 & Patt/ReCoRD\_TH\_drop:default & 77.0 & 640 \\
3 & lex\_glue:case\_hold & 76.64 & 876 \\
4 & sileod/probability\_words\_nli:reasoning\_2\dots & 75.94 & 1258 \\
5 & sst:dictionary & 76.15 & 1162 \\
6 & RussianNLP/russian\_super\_glue:muserc & 74.97 & 1463 \\
7 & go\_emotions:raw & 71.16 & 1551 \\
8 & ltg/norec:default & 74.57 & 1497 \\
9 & metaeval/defeasible-nli:snli & 74.7 & 1489 \\
10 & claudios/cubert\_ETHPy150Open:variable\dots & 76.72 & 814 \\
\bottomrule
\end{tabular}

        \caption{ESM-LogME Ranking: J-STS}
        \label{fig:prob1_6_1}
    \end{minipage}

    \end{tabular}
\end{table}

\begin{table}[H]
    \centering
    \begin{tabular}{cc} 

    \begin{minipage}{0.5\textwidth}
        \centering
        \scriptsize
        \begin{tabular}{l p{4cm} rr}
\toprule
 & Source Task & Perf. & ESM-LM Rank \\
\midrule
1 & kuroneko5943/amz20:Baby & 65.56 & 448 \\
2 & journalists\_questions:plain\_text & 64.65 & 1075 \\
3 & humicroedit:subtask-1 & 64.24 & 969 \\
4 & joelniklaus/lextreme:swiss\_criticality\_pr\dots & 64.03 & 92 \\
5 & evaluate/glue-ci:cola & 64.02 & 257 \\
6 & glue:cola & 64.02 & 389 \\
7 & sbx/superlim-2:dalaj-ged & 63.88 & 241 \\
8 & strombergnlp/nordic\_langid:10k & 63.72 & 1080 \\
9 & kuroneko5943/amz20:CableModem & 63.61 & 54 \\
10 & hpprc/janli:base & 63.49 & 511 \\
\bottomrule
\end{tabular}

        \caption{Ground Truth Ranking: GWQ}
        \label{fig:prob1_6_2}
        \vspace{1cm}
    \end{minipage}%
    \begin{minipage}{0.5\textwidth}
        \centering    
        \scriptsize
\begin{tabular}{l p{4cm} rr}
\toprule
 & Source Task & Perf. & True Rank \\
\midrule
1 & pragmeval:persuasiveness-eloquence & 56.49 & 1144 \\
2 & TheBritishLibrary/blbooksgenre:annotate\dots& 50.76 & 1475 \\
3 & Patt/ReCoRD\_TH\_drop:default & 57.91 & 942 \\
4 & xtreme:PAWS-X.en & 55.41 & 1259 \\
5 & ScandEval/scala-is:default & 59.39 & 596 \\
6 & akhtet/myXNLI:default & 51.83 & 1444 \\
7 & indic\_glue:wstp.mr & 58.54 & 828 \\
8 & sileod/probability\_words\_nli:reasoning\_1\dots & 60.55 & 279 \\
9 & tasksource/mmlu:high\_school\_macroecon\dots & 56.9 & 1082 \\
10 & davebulaval/CSMD:meaning\_holdout\_ide\dots & 58.56 & 821 \\
\bottomrule
\end{tabular}

        \caption{ESM-LogME Ranking: GWQ}
        \label{fig:prob1_6_1}
        \vspace{1cm}
    \end{minipage}
    \end{tabular}
    \vspace{1cm}
    \begin{tabular}{cc}
        \centering
    \begin{minipage}{0.5\textwidth}
        \centering
        \scriptsize
        \begin{tabular}{l p{4cm} rr}
\toprule
 & Source Task & Perf. & ESM-LM Rank \\
\midrule
1 & paws:labeled\_final & 87.4 & 1 \\
2 & xtreme:PAWS-X.en & 87.0 & 234 \\
3 & paws-x:es & 85.7 & 666 \\
4 & paws:unlabeled\_final & 85.4 & 257 \\
5 & paws-x:fr & 85.2 & 317 \\
6 & xtreme:PAWS-X.es & 84.3 & 790 \\
7 & paws-x:de & 84.2 & 812 \\
8 & xtreme:PAWS-X.de & 83.1 & 561 \\
9 & xtreme:PAWS-X.zh & 82.5 & 869 \\
10 & paws-x:zh & 82.5 & 833 \\
\bottomrule
\end{tabular}

        \caption{Ground Truth Ranking: PAWS-X}
        \label{fig:prob1_6_2}
    \end{minipage}%
    \begin{minipage}{0.5\textwidth}
        \centering    
        \scriptsize
\begin{tabular}{l p{4cm} rr}
\toprule
 & Source Task & Perf. & True Rank \\
\midrule
1 & paws:labeled\_final & 87.4 & 1 \\
2 & claritylab/utcd:out-of-domain & 55.4 & 491 \\
3 & tasksource/zero-shot-label-nli:default & 53.8 & 1234 \\
4 & turkish\_product\_reviews:default & 55.3 & 550 \\
5 & swag:full & 53.8 & 1250 \\
6 & go\_emotions:raw & 55.2 & 594 \\
7 & seara/ru\_go\_emotions:raw & 55.2 & 582 \\
8 & davebulaval/CSMD:meaning & 55.6 & 420 \\
9 & metaeval/defeasible-nli:social & 55.5 & 462 \\
10 & TheBritishLibrary/blbooksgenre:annotated\dots & 52.9 & 1470 \\
\bottomrule
\end{tabular}

        \caption{ESM-LogME Ranking: PAWS-X}
        \label{fig:prob1_6_1}
    \end{minipage}
    \end{tabular}
    \vspace{1cm}
    \begin{tabular}{cc}
                \centering
    \begin{minipage}{0.5\textwidth}
        \centering
        \scriptsize
        \begin{tabular}{l p{4cm} rr}
\toprule
 & Source Task & Perf. & ESM-LM Rank \\
\midrule
1 & md\_gender\_bias:opensubtitles\_inferred & 83.0 & 1 \\
2 & md\_gender\_bias:yelp\_inferred & 82.7 & 28 \\
3 & klue:re & 82.5 & 1244 \\
4 & AmazonScience/massive:sw-KE & 82.2 & 1539 \\
5 & AI-Sweden/SuperLim:sweana & 81.7 & 1455 \\
6 & md\_gender\_bias:light\_inferred & 81.6 & 3 \\
7 & DBQ/Mr.Porter.Product.prices.Hungary:de\dots & 81.5 & 1353 \\
8 & conv\_ai\_3:conv\_ai\_3 & 81.4 & 1461 \\
9 & sagteam/author\_profiling:main & 81.3 & 5 \\
10 & DBQ/Gucci.Product.prices.Romania:default & 81.2 & 1336 \\
\bottomrule
\end{tabular}

        \caption{Ground Truth Ranking: MDGB}
        \label{fig:prob1_6_2}
    \end{minipage}%
    \begin{minipage}{0.5\textwidth}
        \centering    
        \scriptsize
\begin{tabular}{l p{4cm} rr}
\toprule
 & Source Task & Perf. & True Rank \\
\midrule
1 & md\_gender\_bias:opensubtitles\_inferred & 83.0 & 1 \\
2 & Patt/ReCoRD\_TH\_drop:default & 69.7 & 1468 \\
3 & md\_gender\_bias:light\_inferred & 81.6 & 6 \\
4 & md\_gender\_bias:wizard & 77.9 & 430 \\
5 & sagteam/author\_profiling:main & 81.3 & 9 \\
6 & art:anli & 73.7 & 1252 \\
7 & md\_gender\_bias:funpedia & 78.1 & 381 \\
8 & omp:posts\_unlabeled & 75.9 & 922 \\
9 & swag:full & 71.2 & 1424 \\
10 & metaeval/defeasible-nli:social & 75.8 & 945 \\
\bottomrule
\end{tabular}

        \caption{ESM-LogME Ranking: MDGB}
        \label{fig:prob1_6_1}
    \end{minipage}
    \end{tabular}
    \vspace{1cm}
    \begin{tabular}{cc}
        \centering
    \begin{minipage}{0.5\textwidth}
        \centering
        \scriptsize
\begin{tabular}{l p{4cm} rr}
\toprule
 & Source Task & Perf. & ESM-LM Rank \\
\midrule
1 & tweet\_eval:offensive & 61.83 & 5 \\
2 & OxAISH-AL-LLM/wiki\_toxic:default & 58.73 & 4 \\
3 & pietrolesci/wikitoxic:default & 58.18 & 3 \\
4 & classla/FRENK-hate-en:multiclass & 56.82 & 8 \\
5 & classla/FRENK-hate-en:binary & 56.18 & 6 \\
6 & hate\_speech\_filipino:default & 56.08 & 106 \\
7 & MichiganNLP/TID-8:md-agreement-atr & 55.47 & 2 \\
8 & MichiganNLP/TID-8:md-agreement-ann & 54.42 & 1 \\
9 & tweet\_eval:emotion & 54.38 & 198 \\
10 & tasksource/crowdflower:text\_emotion & 53.88 & 412 \\
\bottomrule
\end{tabular}

        \caption{Ground Truth Ranking: GCC}
        \label{fig:prob1_6_2}
    \end{minipage}%
    \begin{minipage}{0.5\textwidth}
        \centering    
        \scriptsize
\begin{tabular}{l p{4cm} rr}
\toprule
 & Source Task & Perf. & True Rank \\
\midrule
1 & MichiganNLP/TID-8:md-agreement-ann & 54.42 & 8 \\
2 & MichiganNLP/TID-8:md-agreement-atr & 55.47 & 7 \\
3 & pietrolesci/wikitoxic:default & 58.18 & 3 \\
4 & OxAISH-AL-LLM/wiki\_toxic:default & 58.73 & 2 \\
5 & tweet\_eval:offensive & 61.83 & 1 \\
6 & classla/FRENK-hate-en:binary & 56.18 & 5 \\
7 & sst:dictionary & 50.74 & 30 \\
8 & classla/FRENK-hate-en:multiclass & 56.82 & 4 \\
9 & clue:csl & 47.98 & 88 \\
10 & d0rj/rudetoxifier\_data:default & 49.04 & 59 \\
\bottomrule
\end{tabular}

        \caption{ESM-LogME Ranking: GCC}
        \label{fig:prob1_6_1}
    \end{minipage}

    \end{tabular}
\end{table}

\clearpage
\subsection{Detailed Source Ranking Evaluation}\label{source_ranking_details}
\begin{table}[H]
    \centering
    \begin{tabular}{cc} 

    \begin{minipage}{0.5\textwidth}
        \centering
        \scriptsize


\begin{tabular}{lrrrr}
\toprule
 & NDCG & Regret@1 & Regret@3 & Regret@5 \\
\midrule
IMDB & 78 & 0.74 & 0.74 & 0.74 \\
TEE & 25 & 23.9 & 8.82 & 8.82 \\
TES & 45 & 9.59 & 9.59 & 0 \\
PAWS-X & 63 & 0 & 0 & 0 \\
MDGB & 72 & 0 & 0 & 0 \\
J-STS & 80 & 5.77 & 4.65 & 4.65 \\
GWQ & 59 & 13.83 & 11.66 & 9.41 \\
QCC & 45 & 11.99 & 5.91 & 0 \\
\midrule
avg & 58 & 8.23 & 5.17 & 2.95 \\
\bottomrule
\end{tabular}


        \caption{ESM-LogME Results}
        \label{fig:prob1_6_2}
            \vspace{1cm}
    \end{minipage}%
    \begin{minipage}{0.5\textwidth}
        \centering    
        \scriptsize

\begin{tabular}{lrrrr}
\toprule
 & NDCG & Regret@1 & Regret@3 & Regret@5 \\
\midrule
IMDB & 88 & 0.62 & 0.62 & 0.62 \\
TEE & 77 & 1.84 & 0 & 0 \\
TES & 65 & 10.3 & 0 & 0 \\
PAWS-X & 99 & 0 & 0 & 0 \\
MDGB & 81 & 1.69 & 0 & 0 \\
J-STS & 87 & 2.58 & 2.58 & 2.58 \\
GWQ & 70 & 2.35 & 2.35 & 2.35 \\
QCC & 100 & 0 & 0 & 0 \\
\midrule
avg & 83 & 2.42 & 0.69 & 0.69 \\
\bottomrule
\end{tabular}

        \caption{LogME Results}
        \label{fig:prob1_6_1}
            \vspace{1cm}
    \end{minipage}
    \end{tabular}
    \vspace{1cm}
    \begin{tabular}{cc}
    \begin{minipage}{0.5\textwidth}
        \centering
        \scriptsize
\begin{tabular}{lrrrr}
\toprule
 & NDCG & Regret@1 & Regret@3 & Regret@5 \\
\midrule
IMDB & 73 & 2.96 & 2.1 & 0 \\
TEE & 48 & 6.62 & 1.84 & 1.84 \\
TES & 30 & 5.78 & 3.81 & 3.81 \\
PAWS-X & 84 & 0 & 0 & 0 \\
MDGB & 59 & 6.87 & 4.1 & 3.37 \\
J-STS & 90 & 0 & 0 & 0 \\
GWQ & 57 & 11.74 & 11.74 & 11.74 \\
QCC & 33 & 54.46 & 25.01 & 21.98 \\
\midrule
avg & 59 & 11.05 & 6.07 & 5.34 \\
\bottomrule
\end{tabular}

        \caption{Vocabulary Overlap Results}
        \label{fig:prob1_6_2}
    \end{minipage}%
    \begin{minipage}{0.5\textwidth}
        \centering    
        \scriptsize

\begin{tabular}{lrrrr}
\toprule
 & NDCG & Regret@1 & Regret@3 & Regret@5 \\
\midrule
IMDB & 69 & 6.54 & 4.32 & 2.96 \\
TEE & 39 & 16.18 & 12.5 & 12.5 \\
TES & 27 & 16.5 & 16.5 & 16.5 \\
PAWS-X & 20 & 35.47 & 33.07 & 33.07 \\
MDGB & 74 & 1.69 & 1.69 & 0.36 \\
J-STS & 85 & 3.36 & 3.36 & 3.36 \\
GWQ & 57 & 18.79 & 5.69 & 5.69 \\
QCC & 98 & 0 & 0 & 0 \\
\midrule
avg & 59 & 12.32 & 9.64 & 9.31 \\
\bottomrule
\end{tabular}

        \caption{TaskEmb Results}
        \label{fig:prob1_6_1}
    \end{minipage}
    \end{tabular}
    \vspace{1cm}
    \begin{tabular}{cc}
    \begin{minipage}{0.5\textwidth}
        \centering
        \scriptsize

\begin{tabular}{lrrrr}
\toprule
 & NDCG & Regret@1 & Regret@3 & Regret@5 \\
\midrule
IMDB & 59 & 14.2 & 11.11 & 10.12 \\
TEE & 31 & 18.01 & 18.01 & 18.01 \\
TES & 30 & 3.81 & 3.81 & 3.81 \\
PAWS-X & 83 & 0.46 & 0 & 0 \\
MDGB & 64 & 3.37 & 3.37 & 1.69 \\
J-STS & 89 & 0 & 0 & 0 \\
GWQ & 56 & 21.75 & 4.37 & 4.37 \\
QCC & 18 & 30.21 & 30.21 & 28.94 \\
\midrule
avg & 54 & 11.48 & 8.86 & 8.37 \\
\bottomrule
\end{tabular}

        \caption{TextEmb Results}
        \label{fig:prob1_6_2}
    \end{minipage}%
    \begin{minipage}{0.5\textwidth}
        \centering    
        \scriptsize

\begin{tabular}{lrrrr}
\toprule
 & NDCG & Regret@1 & Regret@3 & Regret@5 \\
\midrule
IMDB & 66 & 3.46 & 0.74 & 0.74 \\
TEE & 65 & 0 & 0 & 0 \\
TES & 15 & 11.42 & 10.01 & 10.01 \\
PAWS-X & 28 & 1.95 & 1.95 & 1.95 \\
MDGB & 58 & 17.71 & 6.27 & 2.29 \\
J-STS & 82 & 5.79 & 2.93 & 2.93 \\
GWQ & 65 & 2.35 & 2.35 & 2.35 \\
QCC & 51 & 17.44 & 0 & 0 \\
\midrule
avg & 54 & 7.51 & 3.03 & 2.53 \\
\bottomrule
\end{tabular}

        \caption{Frozen Transfer Results}
        \label{fig:prob1_6_1}
    \end{minipage}
    \end{tabular}
    \vspace{1cm}
    \begin{tabular}{cc}
    \begin{minipage}{0.5\textwidth}
        \centering
        \scriptsize

\begin{tabular}{lrrrr}
\toprule
 & NDCG & Regret@1 & Regret@3 & Regret@5 \\
\midrule
IMDB & 86 & 0.62 & 0.62 & 0.62 \\
TEE & 52 & 12.87 & 12.87 & 0 \\
TES & 55 & 7.19 & 0 & 0 \\
PAWS-X & 99 & 0 & 0 & 0 \\
MDGB & 78 & 1.69 & 0 & 0 \\
J-STS & - & - & - & - \\
GWQ & - & - & - & - \\
QCC & - & - & - & - \\
\midrule
avg & 74 & 4.47 & 2.7 & 0.12 \\
\bottomrule
\end{tabular}

        \caption{NCE Results}
        \label{fig:prob1_6_2}
    \end{minipage}%
    \begin{minipage}{0.5\textwidth}
        \centering    
        \scriptsize

\begin{tabular}{lrrrr}
\toprule
 & NDCG & Regret@1 & Regret@3 & Regret@5 \\
\midrule
IMDB & 87 & 0.62 & 0.62 & 0.62 \\
TEE & 79 & 0 & 0 & 0 \\
TES & 66 & 7.19 & 0 & 0 \\
PAWS-X & 99 & 0 & 0 & 0 \\
MDGB & 77 & 1.69 & 0 & 0 \\
J-STS & - & - & - & - \\
GWQ & - & - & - & - \\
QCC & - & - & - & - \\
\midrule
avg & 82 & 1.9 & 0.12 & 0.12 \\
\bottomrule
\end{tabular}

        \caption{LEEP Results}
        \label{fig:prob1_6_1}
    \end{minipage}

    \end{tabular}
\end{table}

\clearpage
\section{Datasets}\label{dataset-appendix}
\subsection{Target Datasets}\label{target-datasets-appendix}
\label{sec:appendix}\begin{table}[H]
    \centering

    \begin{minipage}{\textwidth}
\scriptsize
    \centering

        \begin{tabular}{l p{2.4cm}  r p{1cm} p{2cm} p{2.3cm} r}

        \toprule
& Description   
& \# Classes 
& Language
& Config
& Input columns
& Label column\\ \midrule
IMDB
& Sentiment analysis
& 2 
& en
& plain\_text
& text
& label
\\
TES       
& Sentiment analysis                      
& 3
& en
& sentiment
& text
& label
\\
TEE      
& Emotion recognition                     
& 4
& en
& emotion
& text
& label
\\
PAWS-X       
& Paraphrase identification                
& 2
& en
& en
& text
& label
\\
MDGB       
& Gender bias analysis                        
& 2
& en
& convai2\_inferred
& text
& binary\_label
\\
J-STS      
& Semantical similarity
& R
& ja
& JSTS
& sentence1, sentence2
& label
\\
GWQ       
& Query quality analysis                     
& R
& en
& default
& content
& rating
\\
GCC       
& Hate speech detection
& R
& en
& default
& text
& toxicity
\\
        \bottomrule
    \end{tabular}  
            \caption{Target Datasets}

\end{minipage}

    \label{tab:my_label}
\end{table}    



\subsection{Source Datasets}\label{source-datasets-appendix}
The source datasets were heuristically parsed from the Huggingface Hub.
We remove source datasets that are exact duplicates of any of the target datasets. We do not control for duplicates between source datasets.

\begin{figure}[H]
    \centering

\includegraphics[width=0.5\textwidth]{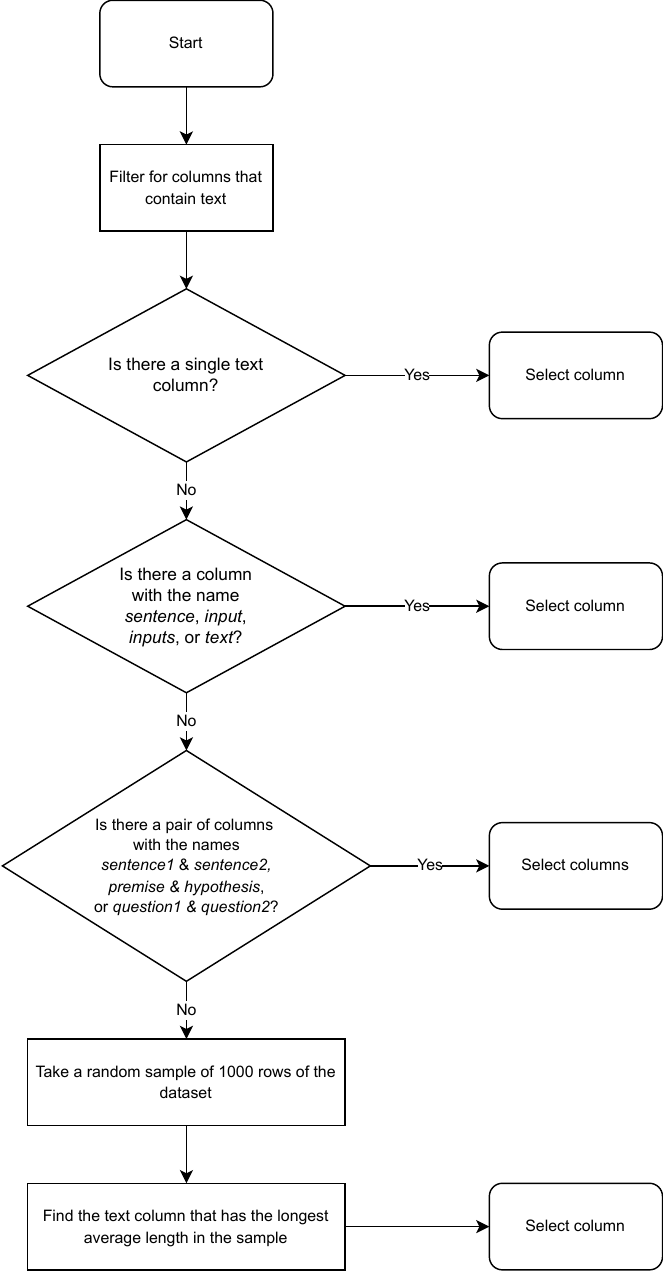}

        \caption{Input Column Assigment}
        \label{fig:prob1_6_2}

\end{figure}
\begin{figure*}
    \centering
\includegraphics[width=0.5\textwidth]{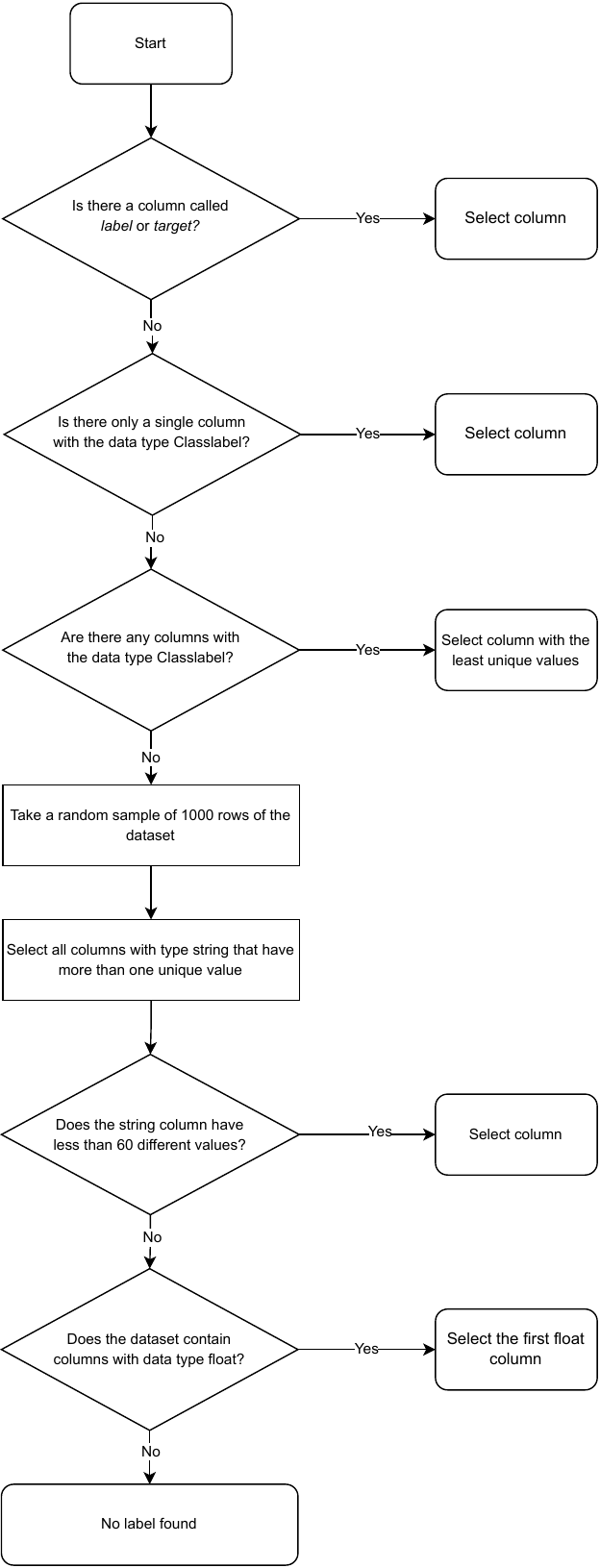}
        \caption{Label Column Assigment}
        \label{fig:prob1_6_1}
\end{figure*}




\end{document}